\documentclass[lettersize,journal]{IEEEtran}
\usepackage{amsmath,amsfonts}
\usepackage{algorithmic}
\usepackage{algorithm}
\usepackage{array}
\usepackage[caption=false,font=normalsize,labelfont=sf,textfont=sf]{subfig}
\usepackage{textcomp}
\usepackage{stfloats}
\usepackage{url}
\usepackage{verbatim}
\usepackage{graphicx}
\usepackage{cite}
\usepackage{bbding}
\usepackage[colorlinks,linkcolor=blue]{hyperref}
\usepackage{amsmath}
\usepackage{multirow}
\usepackage{booktabs}

\begin{document}

\title{Synthesize Boundaries: A Boundary-aware Self-consistent Framework for Weakly Supervised Salient Object Detection}


\author{	\IEEEauthorblockN{Binwei~Xu,
		Haoran~Liang,
		Ronghua~Liang,~\IEEEmembership{Senior Member,~IEEE,}
		and Peng~Chen,~\IEEEmembership{Member,~IEEE}
	}\\
	\thanks{Binwei Xu, Haoran Liang, Ronghua Liang, and Peng Chen are with the College of Computer Science and Technology, Zhejiang University of Technology, Hangzhou 310023, China (e-mail: \{xubinwei, haoran, rhliang, chenpeng\}@zjut.edu.cn). \textit{(Corresponding author: Haoran Liang.)}}
	
}

\markboth{Journal of \LaTeX\ Class Files,~Vol.~14, No.~8, August~2021}%
{Shell \MakeLowercase{\textit{et al.}}: A Sample Article Using IEEEtran.cls for IEEE Journals}


\maketitle

\begin{abstract}
	Fully supervised salient object detection (SOD) has made considerable progress based on expensive and time-consuming data with pixel-wise annotations.
	Recently, to relieve the labeling burden while maintaining performance, some scribble-based SOD methods have been proposed.
	However, learning precise boundary details from scribble annotations that lack edge information is still difficult.
	In this paper, we propose to learn precise boundaries from our designed synthetic images and labels without introducing any extra auxiliary data. The synthetic image creates boundary information by inserting synthetic concave regions that simulate the real concave regions of salient objects.
	Furthermore, we propose a novel self-consistent framework that consists of a global integral branch (GIB) and a boundary-aware branch (BAB) to train a saliency detector. 
	GIB aims to identify integral salient objects, whose input is the original image.
	BAB aims to help predict accurate boundaries, whose input is the synthetic image.
	These two branches are connected through a self-consistent loss to guide the saliency detector to predict precise boundaries while identifying salient objects.
	Experimental results on five benchmarks demonstrate that our method outperforms the state-of-the-art weakly supervised SOD methods and further narrows the gap with the fully supervised methods.
	
\end{abstract}


\section{Introduction}
\IEEEPARstart{S}{alient} object detection (SOD) has rapidly developed and is currently widely applied in many computer vision fields, such as image retrieval~\cite{30he2012mobile,33cheng2017intelligent}, object tracking~\cite{31liang2016adaptive}, and image editing~\cite{32cheng2010repfinder}.
Existing fully supervised SOD methods mostly design different model structures~\cite{8chen2018reverse,20pang2020multi,xu2021locate} or introduce edge-related information~\cite{22zhang2017amulet,23qin2019basnet,24feng2019attentive} to improve performance.
Although considerable progress has been made, these methods rely heavily on pixel-wise annotations, which are expensive and time-consuming to collect.
In recent years, many weakly supervised SOD methods~\cite{wang2017learning, piao2021mfnet, zeng2019multi, zhang2020weakly, yu2021structure} have been explored to relieve the labeling burden while avoiding the degradation of model performance.

\begin{figure}[t]
	\centering
	\includegraphics[width=1\linewidth]{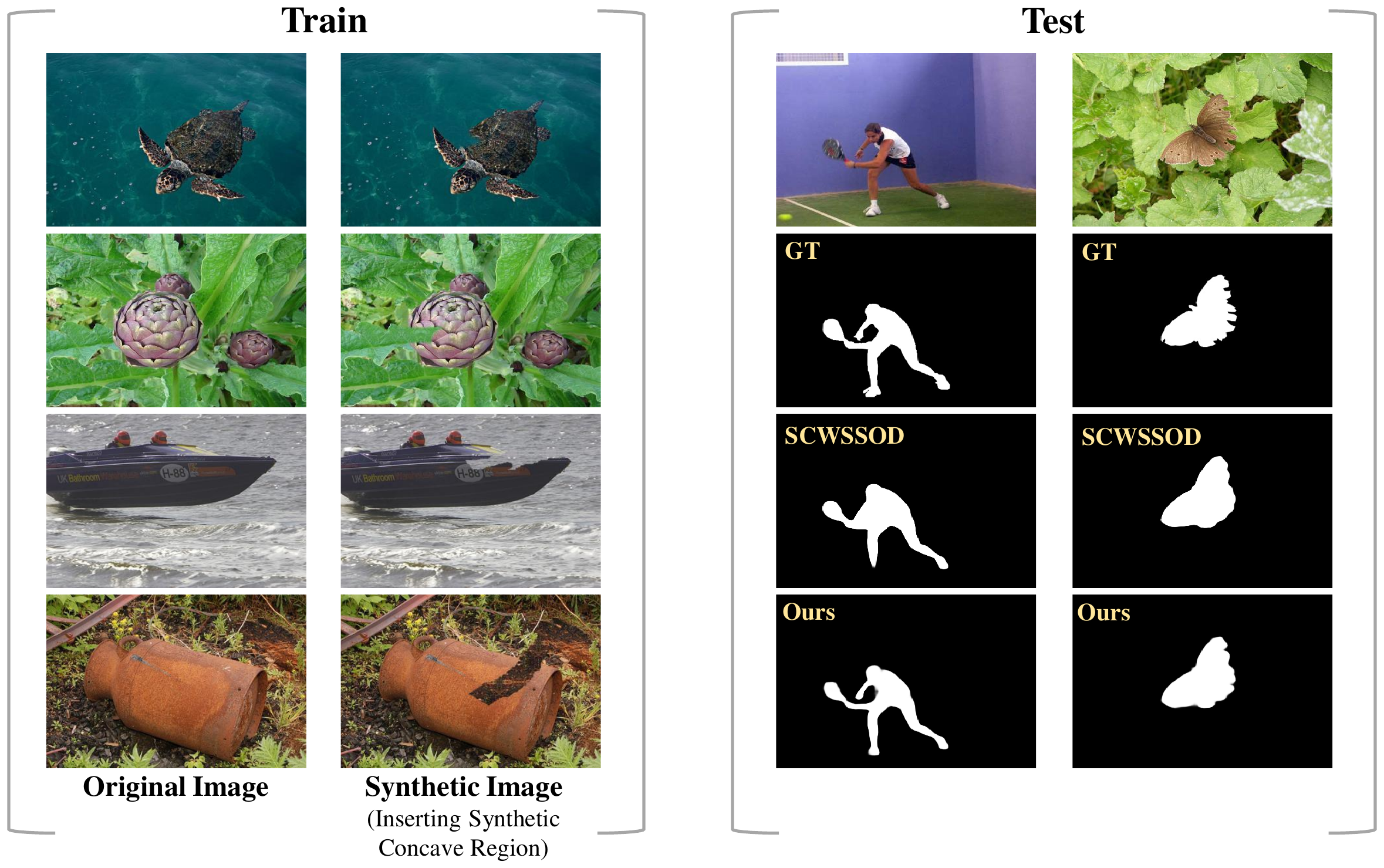} 
	\caption{Sample results of our scribble-based SOD method are compared with SCWSSOD~\cite{yu2021structure}. 
		By adding our proposed synthetic images as training data, our method can perceive tortuous edges and predict a more accurate boundary.
	}
	\label{fig:coverfig}
\end{figure}

User-friendly scribble annotations~\cite{zhang2020weakly} for SOD have been proposed recently. These annotations are located inside the salient objects and the background, and thus, each image can be annotated quickly in several seconds. 
However, models can hardly predict precise boundaries based solely on these sparse scribble annotations.
Therefore,~\cite{zhang2020weakly} introduced an extra edge detection trained on edge datasets to help an SOD model distinguish object boundaries and predict object structure. Nevertheless, this approach introduces an extra trained edge model and does not fundamentally address the problem in which the scribble label itself lacks boundary information.
Furthermore,~\cite{yu2021structure} proposed a structure-consistent weakly supervised salient object detection (SCWSSOD) method that utilizes the intrinsic properties of an image and introduces local saliency coherence (LSC) loss to alleviate the issue of poor edge perception. The core idea is that pixels with close positions and similar RGB values should have similar saliency values. 
Although this method has achieved considerable progress in scribble-based SOD, noisy and incomplete boundary still needs to be optimized.
Just as the tortuous edges or concave region of objects shown in the SCWSSOD results in Fig.~\ref{fig:coverfig}, precisely distinguishing the boundary without the hard pixel-wise label is difficult for a model. 
Besides, for scribble-based SOD task, the partial pixel-wise label are extremely valuable but are shallowly utilized in previous works~\cite{zhang2020weakly, yu2021structure} through partial cross entropy loss without further exploration.

Consequently, to address the problem of poor boundary perception and further explore the scribble labels, we propose to learn salient region discrimination around the boundary from the synthetic concave region (synthetic image of Fig.~\ref{fig:coverfig}).
The fine segmentation of concave region is challenging and also a key for building a better SOD model.
By simulating these concave regions, the synthetic image will obtain their corresponding strong pixel-wise annotations.
These annotations that contains boundary information will drive the model to focus on regions around the edge to produce finer segmentation results.
Here, the synthetic concave region is generated by further exploitation of scribble labels.
Background scribble regions of the image contain background texture and foreground scribble labels can provide local location information of the salient object, which satisfies the conditions of simulating concave regions.

Unlike scribble labels inside the salient objects and the background, synthetic concave regions located at the edge of salient objects can drive the model to focus on the boundary region.
However, the introduction of synthetic concave regions may interfere with a model when identifying salient objects in real images. 
Consequently, we design a self-consistent framework based on real and synthetic images to maintain the model's ability to identify salient objects while predicting precise object boundaries. 
Moreover, the self-consistent framework with self-supervised learning can amplify the advantages of synthetic images to capture more detailed boundary information for better boundary segmentation.
As shown in Fig.~\ref{fig:coverfig}, our method can perceive tortuous edges and concave regions and predict a more accurate boundary.

%

Our main contributions are as follows:
\begin{itemize}
	\item [1.] 
	To the best of our knowledge, we are the first to create boundary supervision for SOD by inserting synthetic concave regions based solely on scribble labels without any extra auxiliary data, alleviating the problem that scribble annotation has no edge information.
	
	\item [2.] 
	A novel self-consistent framework is proposed to amplify the advantages of synthetic images to help a saliency model get more detailed boundaries and maintain the model's ability to identify salient objects. The framework consists of a global integral branch (GIB) that aims to identify integral salient objects and a boundary-aware branch (BAB) that captures precise boundaries.
	
	
	\item [3.] 
	Experimental results demonstrate that our method outperforms state-of-the-art weakly supervised SOD method and is competitive at a certain degree among fully-supervised methods.
	
\end{itemize}

\section{Related Work}
\subsection{Salient Object Detection}
Conventional methods mostly design handcrafted features~\cite{borji2012exploiting,perazzi2012saliency} for SOD. Recently, learning-based approaches have significantly boosted the development of SOD. Most methods capture more accurate regions of salient objects by improving model structure, such as iterative refining~\cite{9deng2018r3net,13wang2019iterative,17wei2019f3net,xu2021locate}, introducing of attention mechanism~\cite{8chen2018reverse,12zhao2019pyramid,14liu2018picanet}, and utilization of efficient fusion strategies~\cite{15zhang2018bi,20pang2020multi,16chen2020global}. 
Some methods~\cite{22zhang2017amulet,23qin2019basnet,24feng2019attentive, zhao2019egnet} use edge-related information that contains rich detailed information to help generate finer segmentation results. Although these methods have achieved considerable progress in SOD, they rely on time-consuming and costly pixel-wise annotations. 

\subsection{Weakly Supervised Salient Object Detection}
With the development of weakly supervised learning, some studies have explored weakly supervised learning in SOD to reduce labeling burden.
\cite{wang2017learning} found that image-level labels can provide foreground information related to salient objects, and they first proposed the use of image-level tags in SOD. 
Then, \cite{piao2021mfnet} found that initial pseudo labels generated from the image-level labels exhibit prejudiced characteristics, so different strategies were proposed to filter out more accurate saliency cues from these noisy pseudo labels for better results.
In addition,~\cite{zeng2019multi} used multimodal sources, including category tags, captions, and noisy pseudo data to train a network by introducing attention transfer loss. 
Besides, a user-friendly annotation method~\cite{zhang2020weakly}, called scribble annotation, was proposed for SOD; this method considerably reduced cost compared with pixel-wise annotation.
Furthermore, ~\cite{yu2021structure} improved the performance of the results by utilizing the intrinsic properties of an image and LSC loss.
Although considerable progress has been achieved, it still demonstrates an inability to predict accurate boundaries of salient objects.
\begin{figure*}[t]
	\centering
	\includegraphics[width=1\textwidth]{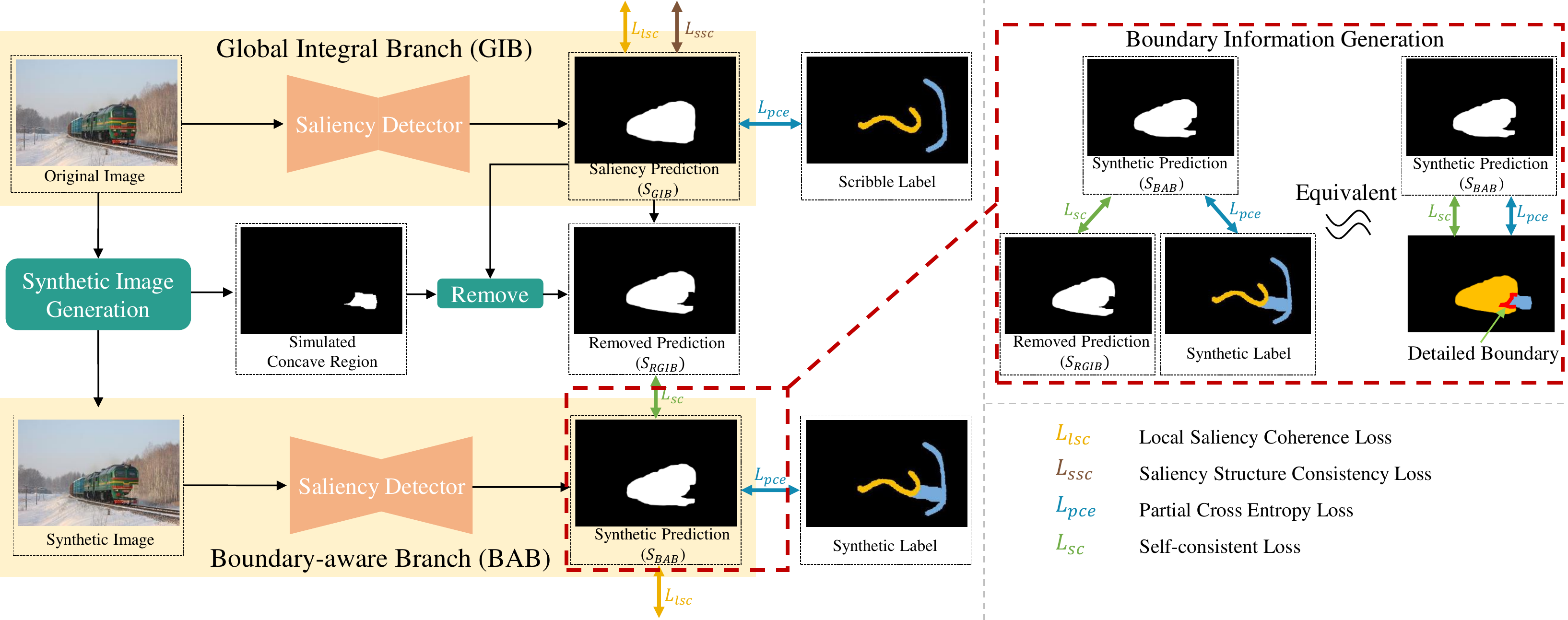} 
	\caption{Overview of our proposed self-consistent framework. It consists of a global integral branch (GIB) and a boundary-aware branch (BAB). GIB trained on original images aims to identify integral salient objects, while BAB trained on synthetic images aims to help predict accurate boundaries. LSC loss $L_{lsc}$ is applied with saliency structure consistency (SSC) loss $L_{ssc}$ and partial cross entropy loss $L_{pce}$ to optimize the global integral branch. Local saliency coherence loss $L_{lsc}$ and partial cross entropy loss $L_{pce}$ are applied to optimize the boundary-aware branch. Self-consistent loss is employed to associate them.}
	\label{fig:framework}
\end{figure*}

\subsection{Scribble Annotation in Similar Field}
Several studies have been conducted on scribble annotations in weakly supervised semantic segmentation and medical image segmentation. ScrribleSup~\cite{lin2016scribblesup} proposed a graphical model for semantic segmentation with scribble annotations. Furthermore, normalized cut loss~\cite{tang2018normalized} and kernel cut loss~\cite{tang2018regularized} were presented to improve the quality of results to be closer to those of fully supervised methods. In addition,~\cite{xu2021scribble} developed a progressive segmentation inference framework through context and annotation inferences to deal with semantic segmentation. Scribble-supervised medical image segmentation focuses on class-specific categories, such as Bohemian glands~\cite{liu2022scribble}, COVID-19 infection~\cite{liu2022weakly}, brain tumors~\cite{ji2019scribble}, and cells~\cite{lee2020scribble2label}. 
Scribble-based semantic segmentation and medical image segmentation have been intensively explored. However, Scribble-based dataset~\cite{zhang2020weakly} is an early-stage dataset that was proposed recently, so little effort has been made so far in scribble-based SOD
and many challenges (e.g., incomplete objects, coarse boundaries, and low accuracy on complex scenes) need to be overcome.

\section{Methodology}

\subsection{Self-consistent Framework}
%

Our self-consistent framework is composed of a global integral branch (GIB) and a boundary-aware branch (BAB) as shown in Fig.~\ref{fig:framework}. 
Specifically, the goal of BAB is to help predict accurate boundaries by driving the saliency detector to focus on boundary part. Its input is the synthetic image generated by inserting the synthetic concave region into the original image.
GIB primarily aims to identify integral salient objects with the supervision of LSC loss, saliency structure consistency (SSC) loss~\cite{yu2021structure}, and partial cross entropy loss to prevent the overfitting of the saliency model trained with synthetic images that gives excessive attention to edge details but disregards the integrity of salient objects.
The self-consistent framework balances these two branches by self-supervised learning to achieve their complementary advantages, \textit{i.e.}, maintaining the model's ability to identify salient objects while predicting precise object boundaries.
The saliency detectors of the two branches are shared-weight siamese networks. 
Here, we adopt the network of~\cite{yu2021structure} as our saliency detector, which uses GCPANet~\cite{chen2020global} with ResNet-50~\cite{he2016deep} backbone pretrained on ImageNet~\cite{deng2009imagenet}.

Besides, the self-consistent framework can amplify the advantages of synthetic images to help BAB capture more detailed boundary information.
Concretely, as shown in boundary information generation (red dotted box) of Fig.~\ref{fig:framework}, the edges of saliency prediction $S_{GIB}$ may be unclear, but salient objects can be detected. Thus, the prediction $S_{RGIB}$ that removes simulated concave region can provide the foreground supervision for BAB via the self-consistent loss. Meanwhile, the simulated concave region of the synthetic label originally provide hard pixel-wise background supervision. 
On the basis of the removed prediction and the synthetic label, more pixel-wise detailed boundary information can be created, driving the network to pay more attention on boundaries.
It is worth noting that even if GIB fails to predict the correct salient object, the simulated concave regions and the foreground scribble regions in the synthetic label are partially adjacent, which still provides partial pixel-wise boundary information.

\subsection{Synthetic Image Generation}
\label{Synthetic_image_part}
Scribble labels are a small number of labeled pixels in the whole image. These pixels are located inside the salient objects and the background. Thus, learning the edge details of salient objects solely via scribble annotations that have no edge-related information is difficult. 
Hence, the network must explore new supervision from different perspectives.
In addition, we find that the concave region of a salient object is one of the parts that are difficult to segment finely. Motivated by the aforementioned consideration and discovery, we design a synthetic image generation method for simulating the real concave region of salient objects. The synthetic concave region is inserted into salient objects, providing edge-related information to alleviate the shortage of the original scribble labels.
Synthetic image generation includes three main steps: endpoint selection, concave region generation, and texture generation.


\subsubsection{Endpoint Selection} 
\begin{figure*}[t]
	\centering
	\includegraphics[width=1\textwidth]{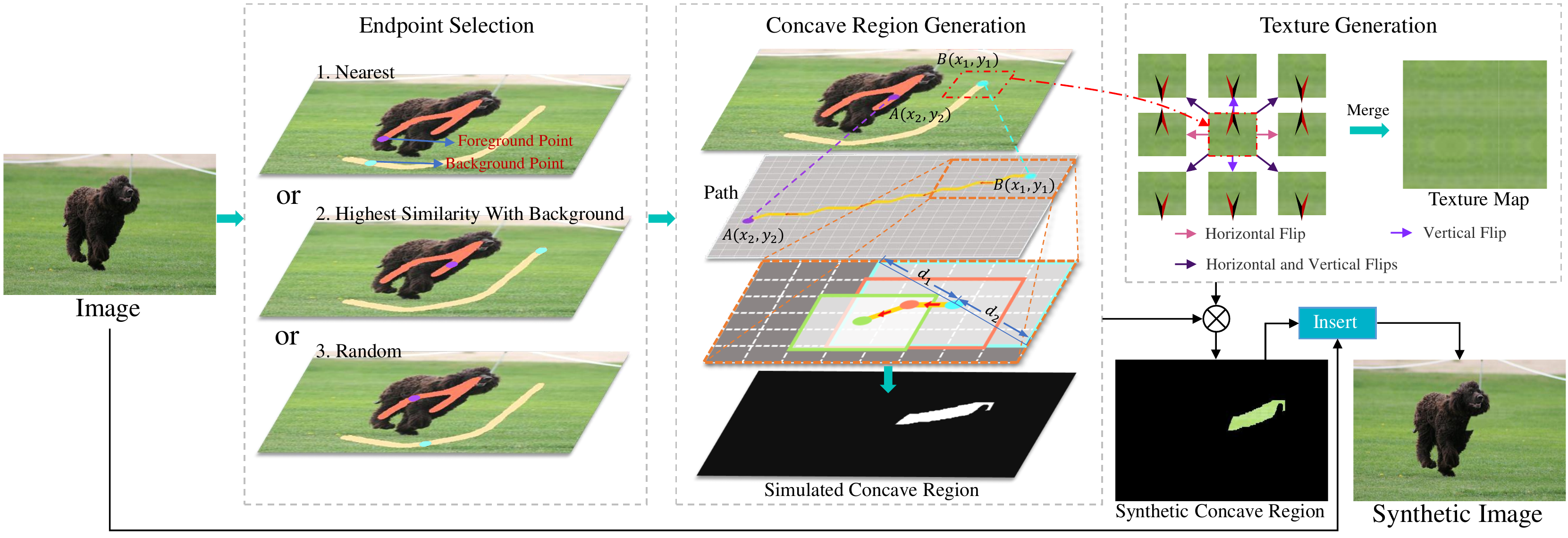} 
	\caption{Illustration of synthetic image generation, which primarily consists of endpoint selection, concave region generation, and texture generation.} 
	\label{fig:synthesis_image}
\end{figure*}

To simulate the real concave region that contains the background information and intersects with salient objects, we select a labeled foreground point and a labeled background point as the endpoints of the synthetic concave region. Concretely, we first perform the skeleton operation on the scribbles before selecting pixels. As shown in Fig.~\ref{fig:synthesis_image}, we design three ways for selecting the foreground and background points. The first one involves finding the two pixels with the closest Euclidean distance as the selected points, ensuring that the shortest path of the simulated concave region and the least change in the original image. However, this solution can only produce one result. To generate more possibilities, we design a second way. We choose the pixel that exhibits the highest similarity with surrounding pixels as the background point and randomly select the foreground point. In this manner, the texture of the generated concave region is more consistent with the surrounding background. Here, similarity to adjacent pixels is calculated by computing the minimum variance in Lab color space in the local $15 \times 15$ window. To create various synthetic backgrounds, the third way randomly selects the background point and foreground point. 
Note that, if other foreground or background points exist on the line between the foreground point and background point, then we choose them as background or foreground points.
In addition, for the first and third ways, we select the point with the highest local similarity to the $15 \times 15$ window around the selected foreground pixel as the final background point to keep the texture of the synthetic concave region as uniform as possible.

\begin{figure}[b]
	\centering
	\includegraphics[width=1\linewidth]{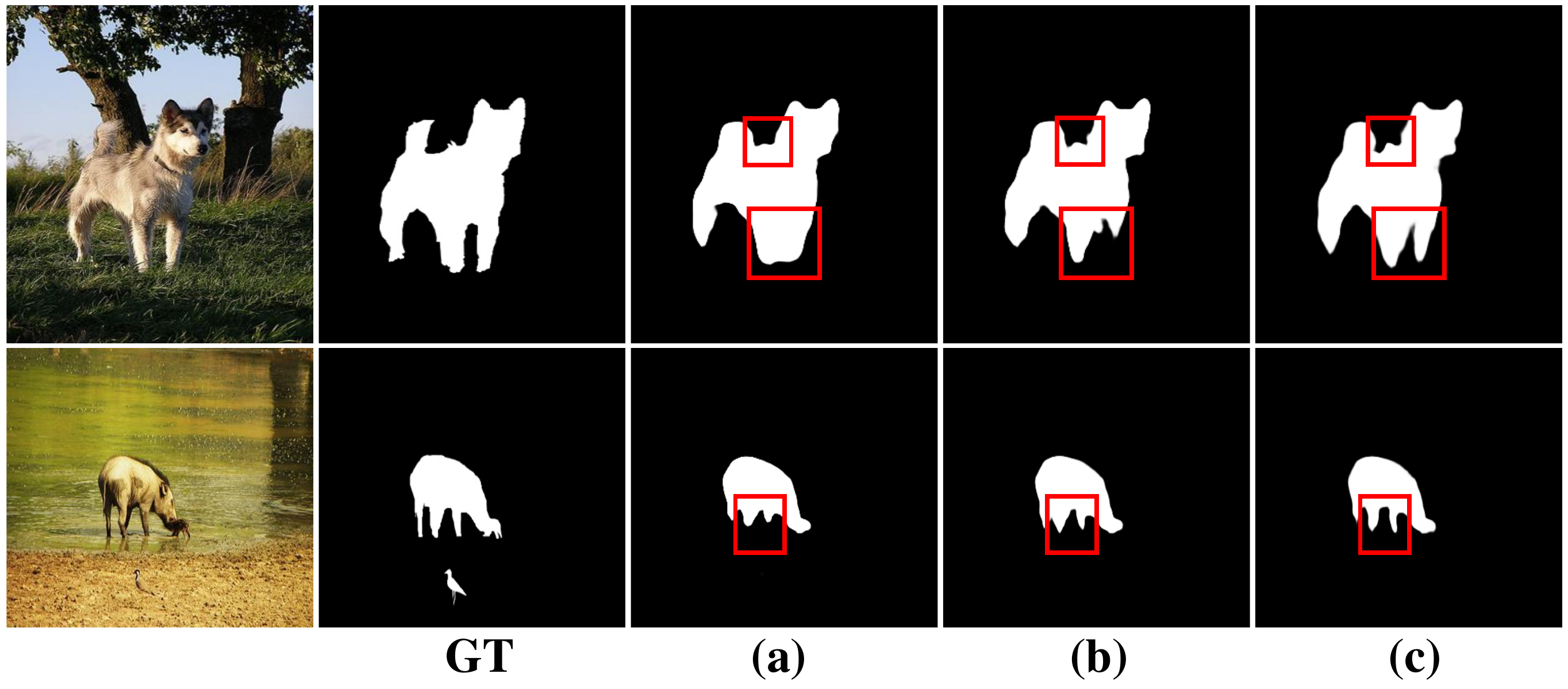} 
	\caption{Comparison of predicted saliency maps for different saliency detectors. (a) a saliency detector without introducing synthetic images as training data. (b) a saliency detector that adds synthetic images as training data. (c) a saliency detector combined with the self-consistent framework.}
	\label{fig:augment_compare}
\end{figure}
\subsubsection{Concave Region Generation} 
Given a background point coordinate $B\left( x_1,y_1  \right)$ and a foreground point coordinate $A\left( x_2,y_2  \right)$, we can generate a path from $B$ to $A$, which is equivalent to the skeleton of the simulated concave region with key information. Specifically, the path generation process starts from $B$ and moves one pixel at a time in the direction of $\left(\dfrac{x_{2}-x_{1}}{\left| x_{2}-x_{1}\right|},0\right)$, $\left( \dfrac{y_{2}-y_{1}}{\left| y_{2}-y_{1}\right|},0\right)$, or $\left(\dfrac{x_{2}-x_{1}}{\left|x_{2}-x_{1}\right|},\dfrac{y_{2}-y_{1}}{\left|y_{2}-y_{1}\right|}\right)$ until it reaches $A$.
After obtaining the path with a length of $N$, we expand the path to generate the simulated concave regions. 
Concretely, we slide the window along the path and then integrate all window regions as the simulated concave region. 
The window size of each path point is unequal. 
We set the basic window size $k \times k$ ($k \in [10, 15]$) and denote the distance $d_1$ from the upper left corner of the window to the $nth$ path point and $d_2$ from the lower right corner to the path point. 
$d_i$ is designed as: 
\begin{equation}
d_{i}=\dfrac{k}{2} \left[ 1+\beta _{1}\dfrac{n}{N} \left( \sin \beta _{2} \dfrac{n}{N} \pi \right) \right], 
\end{equation}
where $i\in \left\{ 0,1\right\}$. $\beta _{1} \in (-1, 1)$ and $\beta _{2} \in (1, 2)$ are randomly set to adjust the shape of simulated concave regions to imitate the shapes of real concave regions. 
After removing the intersection area with the foreground and background, the remainder is the final simulated concave regions.

\subsubsection{Texture Generation} 
To generate the texture map, we crop $15 \times 15$ regions around the background point as the initial local texture. Then, we extend the initial texture through horizontal flips, vertical flips, and horizontal and vertical flips to generate texture maps with any size (as shown in Texture Generation of Fig~\ref{fig:synthesis_image}). 
In this simple way, we can ensure the continuity between textures and preserve the texture of the background. In addition, to make the synthetic background blend into the real background, we set the original local texture position of the generated texture map be the same as the local texture part of the original image.

After obtaining the generated simulated concave regions and texture map, synthetic concave regions with texture can be created by multiplying them.
The final synthetic image is generated by insert it into the original image. Random feathering operation is applied to better incorporate the synthetic concave regions and the original image.

Fig.~\ref{fig:augment_compare} shows the results affected by synthetic image training and self-consistent framework.
We can observe that after the introduction of synthetic images, the edge discrimination of the saliency detector is improved (shown in red box) and the edge of the predicted salient object is more sharpened, which indicates that the synthetic images generated by a simple rule-based method can guide the model to focus on edge. Accordingly, when integrated into our self-consistent framework, the saliency detector can predict more accurate salient boundaries. 
This phenomenon illustrates that our framework can further develop the advantages of synthetic images. 
Meanwhile, the stable and intact prediction suggests that our framework can avoid the degradation of salient object discrimination while enhancing edge segmentation.


\subsection{Loss Function}
As shown in Fig.~\ref{fig:framework}, our loss function consists of three parts: the loss function of GIB $L_{global}$, the loss function of BAB $L_{boundary}$, and the self-consistent loss $L_{sc}$. The training objective $L$ is formulated as: 
\begin{equation}
	L = L_{global} + L_{boundary} + \alpha_1 L_{sc} ,
\end{equation}
where $\alpha_1$ is 0.5.

$L_{global}$ is defined as:
\begin{equation}
	\label{gamma1}
	L_{global}=L_{ssc}+\sum ^{4}_{p=1}\lambda\left( \gamma L^{p}_{lsc}+L^p_{pce}\right),
\end{equation}
where $\lambda$ aims to balance the weight of each decoder stage and we take the same value as SCWSSOD.
$\gamma$ is set to 0.3.
$L_{pce}$ is the partial cross entropy loss, which is defined as:
\begin{equation}
	L_{pce}=\sum _{i\in J}y_{i}\log \widehat{y}_{i}-\left( 1-y_{i}\right) \log \left( 1-\widehat{y}_{i}\right),
\end{equation}
where $y$ represents the ground truth, $\widehat{y}$ is the predicted result, and $J$ is the scribble region with labeled pixels. SSC loss $L_{ssc}$ is applied to enhance generalization ability of our model for different image scales, which can be written as:
\begin{equation}
	L_{ssc} = \alpha_2 \dfrac{1-SSIM\left( S^{\Downarrow },S^{\downarrow } \right) }{2}+\left( 1-\alpha_2 \right) \left| S^{\Downarrow }-S^{\downarrow } \right|,
\end{equation}
where $SSIM$ represents the single scale structural similarity (SSIM)~\cite{wang2004image, godard2017unsupervised} and $\alpha_2$ is set to 0.85. $S^{\downarrow }$ is down-scaled saliency map of a original input image, $S^{\Downarrow }$ is the saliency map of the same image with down-scaled size. LSC loss enforces pixels with similar colors in the local region to share the same saliency label, which
is defined as:
\begin{equation}
	L_{lsc}=\sum _{i}\sum _{j\in K_{i}}F\left( i,j\right) D\left( i,j\right),
\end{equation}
where $D\left( i,j\right) =\left| S_{i}-S_{j}\right|$ is defined as the saliency difference between two pixels $i$ and $j$ and $K_i$ is a local region. $F(i,j)$ is the similarity energy to give close saliency scores for pixels with similar colors and with small distances~\cite{obukhov2019gated}, which is defined as:
\begin{equation}
F\left(i,j\right) =\dfrac{1}{\omega }\exp \left(-\dfrac{\left\| I\left( i\right) -I\left( j\right) \right\| ^{2}}{2\sigma _{I}^{2}}-\dfrac{\left\| P\left( i\right) -P\left( j\right) \right\| ^{2}}{2\sigma _P^{2}}\right).
\end{equation}
$I(\cdot)$ and $P(\cdot)$ are the RGB color and position of a pixel, respectively; $1/{\omega }$ is the normalized weight; $\sigma _{I}$ and $\sigma _P$ are the hyper-parameters of the Gaussian kernel scale; and $\left\| \cdot \right\|$ is an $L2$ operation.

$L_{boundary}$ is defined as: 
\begin{equation}
	\label{gamma2}
	L_{boundary}=\sum ^{4}_{p=1}\lambda\left(\gamma L^{q}_{lsc}+L^q_{pce}\right),
\end{equation}

$L_{sc}$ is combined with SSIM loss, Mean Square Error (MSE), and negative cosine similarity (NCS) loss $L_{ncs}(\cdot, \cdot)$~\cite{chen2021exploring} , which is designed as:

\begin{equation}
	\begin{split}
	L_{sc} = &\alpha_3 \dfrac{1-SSIM\left( S_{RGIB},S_{BAB} \right) }{2}+\\
	&\left( 1-\alpha_3 \right) \left\| S_{RGIB}-S_{BAB} \right\|^2 + \\
	&L_{ncs}(S_{RGIB}, S_{BAB}),
	\end{split}
\end{equation}

where $S_{BAB}$ is the output saliency map of BAB. $S_{RGIB}$ is generated by removing the simulated concave region from the output of GIB. $\alpha_3$ is set to 0.5. 

\begin{table*}[h]
	\caption{Quantitative comparisons with 5 state-of-the-art weakly supervised methods and 8 state-of-the-art fully supervised methods on five datasets in terms of S-measure ($S_m \uparrow$), mean F-measure ($F_\beta \uparrow$), MAE ($M \downarrow$), and E-measure ($E_m \uparrow$). The best results are marked in bold. ``Sup.'' means supervision information. ``F'' means fully supervised. ``I'' means image-level supervised. ``S'' means scribble-level supervised. ``M'' means multi-source supervised.
	}
	\begin{center}
		\setlength{\tabcolsep}{0.5mm}{
			\begin{tabular}{c|c|cccc|cccc|cccc|cccc|cccc}
				\toprule
				& & \multicolumn{4}{c|}{PASCAL-S} & \multicolumn{4}{c|}{ECSSD} & \multicolumn{4}{c|}{HKU-IS} & \multicolumn{4}{c|}{DUT-OMRON} & \multicolumn{4}{c}{DUTS-TE}\\ 
				
				\multicolumn{1}{c}{Methods}
				&\multicolumn{1}{|c|}{Sup.}
				&\multicolumn{1}{c}{$S_m \uparrow$}&{$F_\beta \uparrow$}&{$M \downarrow$}&{$E_m \uparrow$}	
				&\multicolumn{1}{c}{$S_m \uparrow$}&{$F_\beta \uparrow$}&{$M \downarrow$}&{$E_m \uparrow$}
				&\multicolumn{1}{c}{$S_m \uparrow$}&{$F_\beta \uparrow$}&{$M \downarrow$}&{$E_m \uparrow$}	 	 
				&\multicolumn{1}{c}{$S_m \uparrow$}&{$F_\beta \uparrow$}&{$M \downarrow$}&{$E_m \uparrow$}	 
				&\multicolumn{1}{c}{$S_m \uparrow$}&{$F_\beta \uparrow$}&{$M \downarrow$}&{$E_m \uparrow$}	  
				\\ 
				\midrule
				
				DGRL (2018)~\cite{wang2018detect}     & F & .833 & .801 & .073 & .834 & .906 & .903 & .043 & .917 & .897 & .882 & .043 & .941 & .810 & .709 & .063 & .843 & .842 & .794 & .050 &.879\\ 
				MLMSNet (2019)~\cite{wu2019mutual}    & F & .838 & .758 & .073 & .836 & .911 & .869 & .045 & .914 & .906 & .871 & .039 & .953 & .809 & .692 & .064 & .865 & .862 & .745 & .049 &.860\\
				BASNet (2019)~\cite{23qin2019basnet}  & F & .832 & .771 & .075 & .846 & .916 & .880 & .037 & .921 & .909 & .895 & .032 & .946 & .836 & .756 & .056 & .869 & .866 & .791 & .048 &.884\\ 
				PoolNet (2019)~\cite{liu2019simple}   & F & .843 & .815 & .074 & .848 & .921 & .915 & .039 & .924 & .916 & .899 & .033 & .948 & .836 & .747 & .056 & .863 & .883 & .809 & .040 &.889\\
				MINet (2020)~\cite{20pang2020multi}   & F & .850 & .829 & .063 & .851 & .925 & .924 & .033 & .927 & .919 & .909 & .029 & .953 & .833 & .755 & .055 & .865 & .884 & .828 & .037 &.898\\ 
				GateNet (2020)~\cite{zhao2020suppress}& F & .851 & .819 & .067 & .851 & .920 & .916 & .040 & .924 & .915 & .899 & .033 & .949 & .838 & .746 & .055 & .862 & .885 & .807 & .040 &.889\\
				GCPANet (2020)~\cite{16chen2020global}& F & \textbf{.858} & .827 & \textbf{.061} & .847 & .927 & .919 & .035 & .920 & .920 & .898 & .031 & .949 & .839 & .748 & .056 & .860 & .891 & .817 & .038 & .891\\ 
				SGLKRN (2021)~\cite{xu2021locate}     & F & .849 & .830 & .067 & \textbf{.859} & .923 & .922 & .036 & .927 & \textbf{.921} & \textbf{.916} & \textbf{.028} & \textbf{.954} & \textbf{.846} & \textbf{.783} & \textbf{.049} & \textbf{.883} & \textbf{.893} & \textbf{.851} & \textbf{.034} &\textbf{.913}\\
				ICON-R (2022)~\cite{zhuge2022salient} & F & .855 & \textbf{.833} & .063 & .855 & \textbf{.929} & \textbf{.928} & \textbf{.032} & \textbf{.929} & .920 & .910 & .029 & .952 & .844 & .772 & .057 & .870 & .889 & .838 & .037 & .902\\
				\midrule
				
				MWS (2019)~\cite{zeng2019multi}       & M & .761 & .712 & .132 & .784 & .827 & .840 & .096 & .884 & .818 & .814 & .084 & .895 & .756 & .609 & .109 & .763 & .759 & .685 & .091 &.814\\ 
				MFNet (2021)~\cite{piao2021mfnet}     & I & .775 & .746 & .111 & .812 & .837 & .844 & .084 & .887 & .852 & .839 & .058 & .917 & .726 & .621 & .098 & .783 & .778 & .693 & .079 &.830\\
				NSAL (2022)~\cite{piao2022noise}      & I & .761 & .756 & .110 & .816 & .834 & .856 & .078 & .883 & .854 & .864 & .051 & .923 & .745 & .648 & .088 & .801 & .782 & .730 & .073 &.849\\ 
				WSSA (2020)~\cite{zhang2020weakly}    & S & .791 & .774 & .092 & .831 & .865 & .870 & .059 & .901 & .865 & .860 & .047 & .927 & .785 & .703 & .068 & .840 & .803 & .742 & .062 &.857\\
				SCWSSOD (2021)~\cite{yu2021structure} & S & .813 & .818 & .077 & .846 & .882 & .900 & .049 & .908 & .882 & .896 & .038 & .938 & .812 & .758 & \textbf{.060} & .862 & .841 & .823 & .049 &.890\\ 
				Ours                                  & S & \textbf{.825} & \textbf{.827} & \textbf{.073} & \textbf{.853} & \textbf{.890} & \textbf{.906} & \textbf{.046}  & \textbf{.917} & \textbf{.895} & \textbf{.906} & \textbf{.034} & \textbf{.949} & \textbf{.818} & \textbf{.762} & \textbf{.060} & \textbf{.868} & \textbf{.853} & \textbf{.834} & \textbf{.045} & \textbf{.901}\\
				\bottomrule

		\end{tabular}}
	\end{center}
	
	\label{table:state_of_the_art}
\end{table*}

\begin{figure*}[t!]
	\centering
	\includegraphics[width=1\textwidth]{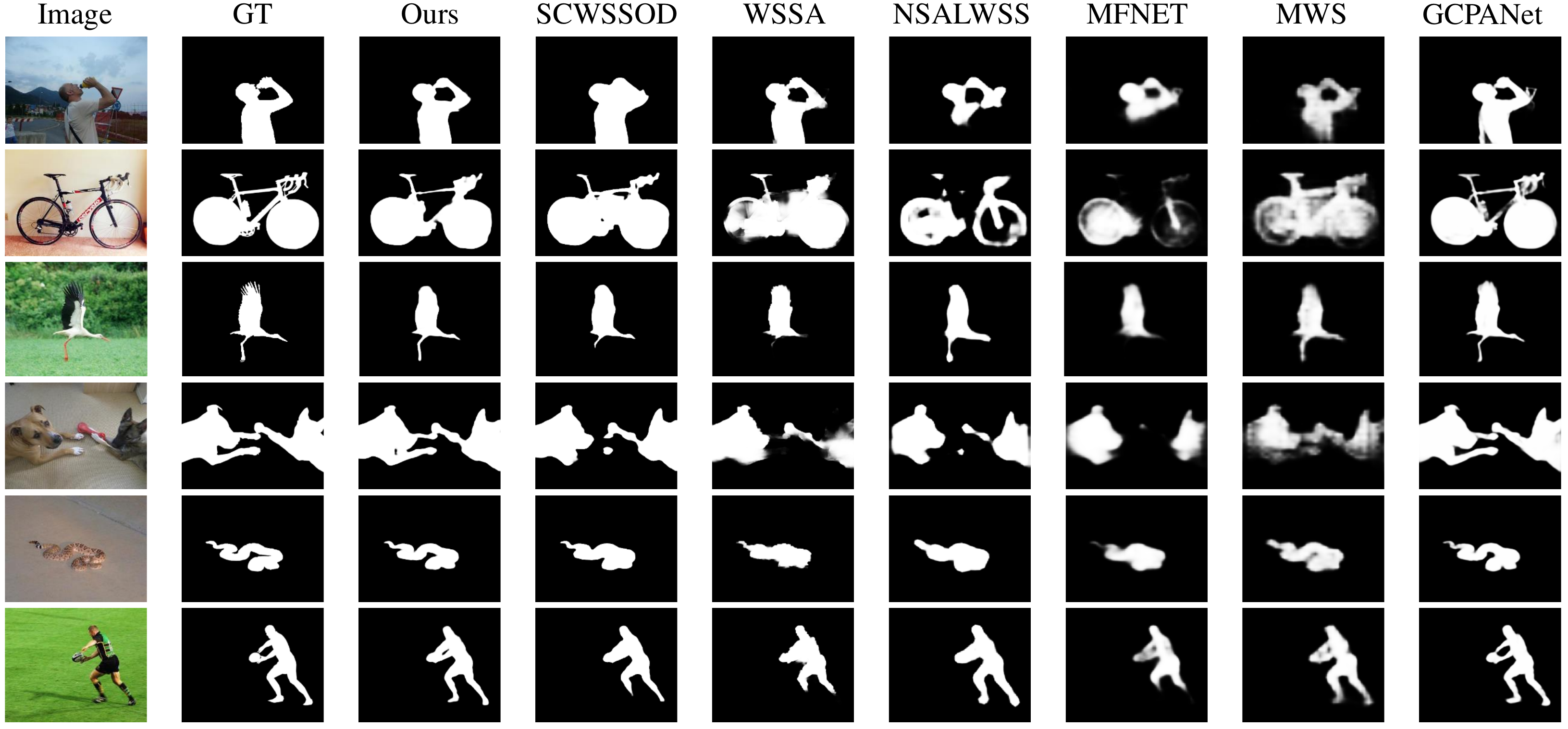} 
	\caption{Visual comparisons of various method. Each column denotes one approach and each row demonstrates saliency maps of one image. Apparently, our method can predict more complete salient objects and can more clearly distinguish the boundaries of
		salient objects than other methods.}
	\label{fig:visual_map}
\end{figure*}

\begin{figure*}[t!]
	\centering
	\includegraphics[width=1\textwidth]{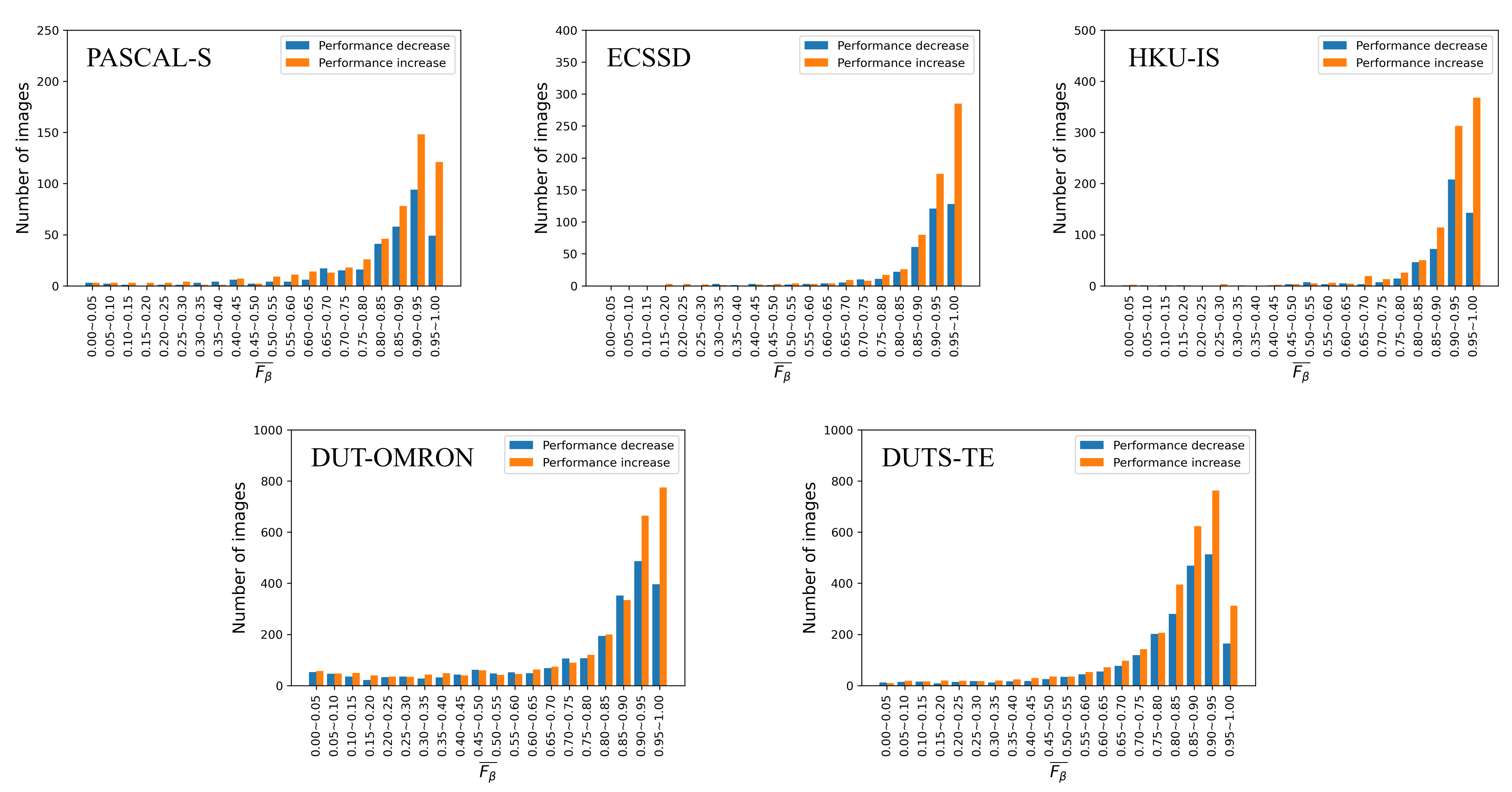} 
	\caption{A visual graph that shows the number of performance increases and decreases of our method compared with the baseline on images with different detection difficulties on five public datasets. Average F-measure $\overline{F_{\beta }}$ of our method and the baseline as an indicator of detection difficulty.}
	\label{fig:analize_final}
\end{figure*}

\section{Experiments}

\subsection{Implementation Details and Datasets}
\subsubsection{Implementation Details}
We adopt stochastic gradient descent (SGD) optimizer with a batch size of 16, momentum of 0.9, and a weight decay of 5e-4. In addition, triangular warm-up and decay strategies with the minimum learning rate of 1e-5 and the maximum learning rate of 5e-3 are used to train our model with 55 epochs.
Each input image is resized to 320$\times$320 and we use horizontal flips and random rotation for data augmentation.
All experiments are run on an NVIDIA GeForce GTX 3090.
It's time-consuming to generate synthetic images online during training, so we generate 10 synthetic images of each original image before training and randomly select one of these synthetic images during training.
To get complete salient objects and then optimize the boundary, during training, we train the global integral branch in the first 28 epochs and then train the whole model.

\subsubsection{Datasets and Evaluation Metrics} 
We train our model on scribble annotated dataset S-DUTS~\cite{zhang2020weakly}. To evaluate the performance of our proposed method, we implement experiments on five public benchmark datasets: ECSSD~\cite{2yan2013hierarchical}, DUT-OMRON~\cite{yang2013saliency}, HKU-IS~\cite{li2015visual}, DUTS-TE~\cite{wang2017learning}, and PASCAL-S~\cite{li2014secrets}.
We adopt four widely-used metrics in our experiments: mean F-measure ($F_\beta$)~\cite{achanta2009frequency}, structure-based metric ($S_m$)~\cite{fan2017structure}, mean absolute error (MAE), and E-measure ($E_m$)~\cite{fan2018enhanced}.

\subsection{Comparison with the State-of-the-arts}
\subsubsection{Quantitative Comparison.}
We compare our method with 5 state-of-the-art weakly supervised SOD methods (SCWSSOD~\cite{yu2021structure}, WSSA~\cite{zhang2020weakly}, NSAL~\cite{piao2022noise}, MFNet~\cite{piao2021mfnet}, MWS~\cite{zeng2019multi}) and 9 fully supervised SOD methods (ICON-R~\cite{zhuge2022salient}, SGLKRN~\cite{xu2021locate}, GCPANet~\cite{16chen2020global}, GateNet~\cite{zhao2020suppress}, MINet~\cite{20pang2020multi}, PoolNet~\cite{liu2019simple}, BASNet~\cite{23qin2019basnet}, MLMSNet~\cite{wu2019mutual}, DGRL~\cite{wang2018detect}). For a fair comparison, the saliency maps of other methods are provided by authors or generated by the released codes and we evaluate them with the same evaluation code.
The best results are bold. As shown in Table~\ref{table:state_of_the_art}, our method obviously outperforms state-of-the-art weakly supervised SOD methods. Compared with the previous best scribble-based SOD method (SCWSSOD), our method performs better on all datasets.
These results demonstrate that our proposed method is effective and robust.
What’s more, our method is comparable or even superior to some fully supervised methods, such as DGRL, BASNet, and MLMSNet.
For some metrics such as $F_\beta$ and $E_m$, our method performs better compared to most fully supervised methods on PASCAL-S, HKU-IS, DUT-OMRON, and DUTS-TE.
We devise a simple rule-based method with certain limitations, such as repeated textures, to simulate concave regions, but does not prevent good results of the saliency detector.
This just proves the reliability of our idea of creating boundary information by simulating concave regions and demonstrates that our self-consistent framework can effectively help synthetic concave regions to realize its value.


\subsubsection{Qualitative Comparison}
To further valuate the advantages of our method, we provide visual examples of the proposed approach and other state-of-the-art methods in Fig.~\ref{fig:visual_map}.
It can be seen that our method can predict more precise and complete results compared with state-of-the-art weakly supervised methods and is close to the supervised method (GCPANet). Rows 1, 2, and 3 demonstrate that our model can distinguish foreground and background and capture the intact salient objects. Rows 3, 4, 5, and 6 demonstrate the ability of our method to predict accurate boundaries.

\subsection{Ablation Study}

\subsubsection{Effectiveness of Synthetic Images}
To prove the effectiveness of synthetic images, we directly use them as a part of the training data to train a saliency detector.
Table~\ref{table:ablation} shows the results of the saliency detector with and without adding synthetic images as training data. 
We adopt a single GIB with the SSC loss, LSC loss, and partial cross entropy loss as the baseline model.
``Syn.'' means whether synthetic images are employed as a part of the training data. ``w/o'' means that the model is trained on original images and ``w/'' means that the model is trained on original images and synthetic images.
It can be seen that the results of the baseline model are improved when the synthetic images are added, which proves that our idea of capturing boundary information by simulating concave regions is reliable and the technical solution of generating synthetic concave regions based on a simple rule-based way is practical.

\begin{table}
	\caption{Ablation study for our synthetic images and self-consistent framework. ``SCF'' denotes for self-consistent framework.}
	\centering

	\setlength{\tabcolsep}{0.7mm}{
		\begin{center}
			\begin{tabular}{c|c|cccc|cccc}
				\toprule
				\multirow{2}*{Method} & \multirow{2}*{Syn.}  & \multicolumn{4}{c|} {DUTS-TE} & \multicolumn{4}{c}{PASCAL-S}   \\
				&&$S_m \uparrow$ &$F_\beta \uparrow$ &$M \downarrow$ &$E_m \uparrow$ &$S_m \uparrow$ &$F_\beta \uparrow$ &$M \downarrow$ &$E_m \uparrow$\\
				\midrule
				Baseline&  w/o &                        .842 & .814 & .049 & .889  &.814 & .815 &.079 & .847\\
				Baseline& w/&                           .845 & .820 & .048 & .893  &.816 & .818 &.078 & .849\\
				Baseline+SCF& w/ &                   .853 & .834 & .045    & .901  &.825 & .827 &.073 & .853\\
				
				\bottomrule
				
			\end{tabular}
	\end{center}}
	
	\label{table:ablation}
\end{table}

\subsubsection{Effectiveness of Self-consistent Framework}
To evaluate the effectiveness of the whole framework, we compare the results of the baseline model trained only on the original images and those of introducing the self-consistent framework.
As shown in Table~\ref{table:ablation}, when the baseline model is combined with our framework, the performance is significantly improved, which proves that our proposed framework combined with synthetic images can strengthen the saliency model.
In addition, we respectively replace the saliency detector with three popular SOD models (\textit{i.e.}, PoolNet, MINet, and GateNet) to further validate our framework.
We compare the baseline results of these three models with and without introducing the self-consistent framework.  
For a fair comparison, the configuration is the same for each group of comparisons. 
As shown in~Table~\ref{table:framework_ablation}, all models with our framework perform significantly better, which further demonstrates that our proposed framework is reliable, practical, and flexible. 

Furthermore, to prove that our framework can effectively help the synthetic image amplify its advantages while maintaining the model's ability to accurately identify salient objects, we compare the results of the baseline model that introduces synthetic images directly as a part of training data with those of our self-consistent framework. 
As shown in Table~\ref{table:ablation}, compared with only using synthetic images for data augmentation, integrating synthetic images into our self-consistent framework can achieve significant improvement.

\begin{table}
	\centering
	\caption{Quantitative comparison of different models with and without the self-consistent framework (SCF).}
	\setlength{\tabcolsep}{0.7mm}{
		\begin{tabular}{c|c|cccc|cccc}
			\toprule
			\multirow{2}*{Method}& \multirow{2}*{SCF}  & \multicolumn{4}{c|} {DUTS-TE} & \multicolumn{4}{c}{PASCAL-S}   \\
			&&$S_m \uparrow$ &$F_\beta \uparrow$ &$M \downarrow$ &$E_m \uparrow$  &$S_m \uparrow$ &$F_\beta \uparrow$ &$M \downarrow$ &$E_m \uparrow$\\
			\midrule
			
			\multirow{2}{*}{PoolNet}    &w/o& .824 & .755 & .059 & .870
			& .812 & .743 & .085 & .833\\  
			&w& .837 & .766 & .053 & .881
			& .820 & .753 & .079 & .840\\ \midrule
			\multirow{2}{*}{MINet} &w/o& .833 & .763 & .055 & .877
			& .819 & .761 & .076 & .842\\ 
			&w& .844 & .772 & .051 & .884
			& .826 & .764 & .073 & .844\\ \midrule
			\multirow{2}{*}{GateNet} &w/o& .832 & .781 & .056 & .873
			& .810 & .792 & .081 & .818\\ 
			&w& .845 & .800 & .050 & .891
			& .817 & .792 & .076 & .845\\  
			
			\bottomrule
	\end{tabular}}
	
	\label{table:framework_ablation}
\end{table}

\subsubsection{Visual Analysis of Results}




The main idea of our method is to help the model get more detailed boundaries while maintaining the model's ability to identify salient objects.
Therefore, we create a visual graph to analyze how results are improved and whether our approach of introducing synthetic images leads to a decrease in the model's ability to identify salient objects.
As shown in Fig.~\ref{fig:analize_final}, we count the number of performance increases and decreases of our method compared with the baseline on images with different detection difficulties.
We believe that images that perform well on both our method and the baseline are easy to detect, and vice versa, so we use the average F-measure $\overline{F_{\beta }}$ of our method and the baseline as an indicator of detection difficulty.
It can be seen that the total number of increases is much more than the number of decreases, which indicates that our model solidly improves performance on most images rather than mostly bigger gains on fewer images.
In addition, for relatively easy images, especially for $\overline{F_{\beta }}$ of 0.90 to 1.00, our method improves performance on most images.
The reason is that salient objects can be easily recognized in images with high $\overline{F_{\beta }}$. Based on the accurate and robust identification of the complete salient objects, our method that further refines the boundaries leads to a valid improvement in results.
But for images with low $\overline{F_{\beta }}$, the model usually misrecognizes the salient objects or does not identify the complete structure of the salient objects. 
Regarding these images, even after the same training process, the same model will predict different results each time and the difference is likely to exceed the improvement of the boundary by our method.
In other words, compared with the edge details, the body of the salient objects dominates the result for these images.
Thus, the number of performance increases is not exactly more than the number of performance decreases on hard-to-detect images.
But from another perspective, similar or slightly better performance on these difficult images illustrates that our method does not degrade the discrimination ability to salient objects.

\subsubsection{Comparison with Random Erasing}
Unlike simple data augmentations, we focus on treating a specific problem of the poor handling of concave regions and protruding elements such as animal limbs, and design specific synthetic images as training data augmentations to deal with it.
To demonstrate the effectiveness of the synthetic images and prove that our results are not improved by data extensions, we compare the results of our method with those based on simple random erasing~\cite{zhong2020random}.
For fair comparisons, we directly replace synthetic images with the images after random erasing.
As shown in Table.~\ref{table:erasing}, using random erasing as data augmentation for the baseline even slightly degrades the performance of the model.
In addition, compared with synthetic images, introducing random erasing into the self-consistent framework does not significantly improve the performance. 
These results demonstrate that the model is difficult to be enhanced by simply enriching the data and our custom-designed synthetic images for boundary enhancement are the key to strengthening the model. 

\begin{table}[t]
	\centering
	\caption{Ablation study for our method based on synthetic concave region and those based on random erasing. ``SCF'' denotes for self-consistent framework, ``RE'' denotes for random erasing and ``Syn'' denotes for synthetic images.}
	\setlength{\tabcolsep}{1mm}{
		\begin{tabular}{c|cccc}
			\toprule
			$k$ &$S_m \uparrow$ &$F_\beta \uparrow$ &$M \downarrow$ &$E_m \uparrow$\\
			\midrule
			Baseline&             .842 & .814 & .049 & .889 \\
			Baseline + RE&        .840 & .811 & .049 & .886 \\
			Baseline + Syn&		  .845 & .820 & .048 & .893 \\
			Baseline + SCF + RE & .844 & .817 & .048 & .891 \\
			Baseline + SCF + Syn& .853 & .834 & .045 & .901 \\
			\bottomrule
			
	\end{tabular}}
	
	\label{table:erasing}
\end{table}

\begin{table}
	\centering
	\caption{Ablation analysis for different consistent losses.}
	\setlength{\tabcolsep}{1mm}{
		\begin{tabular}{ccc|cccc}
			\toprule
			MSE& SSIM & NCS &$S_m \uparrow$ &$F_\beta \uparrow$ &$M \downarrow$ &$E_m \uparrow$ \\
			\midrule
			$\surd$&&&                  		.840   & .820    & .048  &.892 \\
			$\surd$&$\surd$&&                  .848   & .828    & .046  &.896 \\
			$\surd$&$\surd$&$\surd$&          .853 & .834 & .045 &.901\\
			\bottomrule
			
	\end{tabular}}
	
	\label{table:sc_loss}
\end{table}

\begin{table}[t]
	\centering
	\caption{Ablation analysis for different widths of synthetic concave regions.}
	\setlength{\tabcolsep}{1mm}{
		\begin{tabular}{c|cccc}
			\toprule
			$k$ &$S_m \uparrow$ &$F_\beta \uparrow$ &$M \downarrow$ &$E_m \uparrow$\\
			\midrule
			$5\sim10$&                  .850 & .830 & .045 & .899 \\
			$10\sim15$&                 .853 & .834 & .045 & .901 \\
			$15\sim20$&                 .848 & .827 & .046 & .897 \\
			\bottomrule
			
	\end{tabular}}
	
	\label{table:kernel}
\end{table}
\subsubsection{Different Self-consistent Losses} 
As shown in Table~\ref{table:sc_loss}, we evaluate the results of different self-consistent losses on DUT-TE dataset. It can be seen that our method combining MSE, SSIM loss, and NCS loss performs best.

\subsubsection{Different Widths of Synthetic Concave Regions} Basic window size $k$ is randomly chosen from a fixed numerical range to adjust the width of every synthetic concave region. As shown in Table \ref{table:kernel}, we compare the results of three different numerical ranges on DUT-TE dataset and the model performs best when $k$ is set from 10 to 15.

\section{Conclusion}
To alleviate the issue of poor model performance caused by scribble labels without edge information, we first propose to create boundary information by designing synthetic images without introducing any extra data. 
Besides, we present a novel self-consistent framework to capture more detailed boundary information for predicting precise object boundaries.
Extensive experiments well demonstrate that the proposed method can perceive tortuous edges and predict fine boundaries. 
It is worth noting that we achieve good results by simply setting the shape and texture of synthetic concave regions in a rule-based way, which proves our core idea of creating boundary information by simulating the real concave region is reliable and effective. 
For future work, we will explore a learning-based approach to improve the rule-based method and apply the proposed method to solve a wide range of weakly supervised tasks with scribble labels in computer vision.
Although considerable progress has been made, we clearly know the limitation of our rule-based synthetic images. 
Therefore, we are going to develop synthetic methods by combining the generative adversarial network.

\section*{Acknowledgment}
This work was partially supported by the National Key Research and Development Program of China (2020YFB1707700), the National Natural Science Foundation of China (62176235, 62036009, 61871350, U1909203), and Zhejiang Provincial Natural Science Foundation of China (LY21F020026).

{
	\bibliographystyle{IEEEtran}
	\bibliography{egbib}
}

\newpage

\vfill

\end{document}